\documentclass[10pt,conference,a4paper]{IEEEtran}
\usepackage{graphicx}
\usepackage{amsmath}
\usepackage{booktabs}

\begin{document}

\title{Retinal Cyst Detection from Optical Coherence Tomography Images}

\author{
    \IEEEauthorblockN{
        Abhishek Dharmaratnakar, 
        Aadheeshwar Vijayakumar, 
        Suchand Dayanand
    }
    \IEEEauthorblockA{
        \textit{Department of Computer Science and Engineering} \\
        \textit{National Institute of Technology Karnataka, Surathkal}\\
        Mangaluru, India \\
    }
}

\maketitle

\begin{abstract}
Retinal Cysts are formed by leakage and accumulation of fluid in the retina due to the incompetence of retinal vasculature. These cystic spaces have significance in several ocular diseases such as age-related macular degeneration, diabetic macular edema, etc. Optical coherence tomography is one of the predominant diagnosing techniques for imaging retinal pathologies. Segmenting and quantification of intraretinal cysts plays the vital role in predicting visual acuity. In literature, several methods have been proposed for automatic segmentation of intraretinal cysts. As cystoid macular edema becomes a major problem to humankind, we need to quantify it accurately and operate it out, else it might cause many problems later on. Though research is being carried out in this area, not much of progress has been made and accuracy achieved so far is 68\% which is very less. Also, the methods depend on the quality of the image and give very low results for high noise images like topcon. This work uses ResNet CNN (Convolutional Neural Network) approach of segmentation by the way of patchwise classification for training on image set from cyst segmentation challenge dataset and testing on test data set given by 2 different graders for all 4 vendors in the challenge. It also compares these methods using first publicly available novel cyst segmentation challenge dataset. The methods were evaluated using quantitative measures to assess their robustness against the challenges of intraretinal cyst segmentation. The results are found to be better than the previous state of the art approaches giving more than 70\% dice coefficient on all vendors irrespective of their quality.
\end{abstract}

\section{Introduction}
Cystoid macular edema (CME) is a pathological consequence of several ocular disorders including diabetic retinopathy, retinal vein occlusion, ocular inflammation, and age-related macular degeneration [1]. Diabetic retinopathy and age-related macular degeneration are leading causes of irreversible blindness in the U.S.. The number of people expected to experience vision loss is predicted to double over the next 30 years. The presence of CME in these conditions is often associated with loss of visual acuity. New and improved methods are needed for the identification and characterization of CME to enhance prevention and inform treatment options for vision loss [11].

Optical coherence tomography (OCT) depth resolves optical reflections from internal structures in biological tissues by using noninvasive, low-coherence light [6]. OCT is widely employed for the assessment of macular diseases and has enabled detailed characterizations of CME. As shown in Fig. 1, OCT is highly effective for visualizing CME because the cystoid fluid has less optical scattering than the surrounding retinal tissues. Typical methods for OCT-based assessment in disorders associated with CME involve the measurement of foveal thickness because of its strong anticorrelation with visual acuity. However, a recent study describes CME in the absence of macular thickening in several retinal disorders and recognizes that CME may not always be associated with macular thickening [10]. Measurements of macular thickness can also be more error prone in the presence of subretinal fluid.

\begin{figure}[htbp]
\centering
\includegraphics[width=\linewidth]{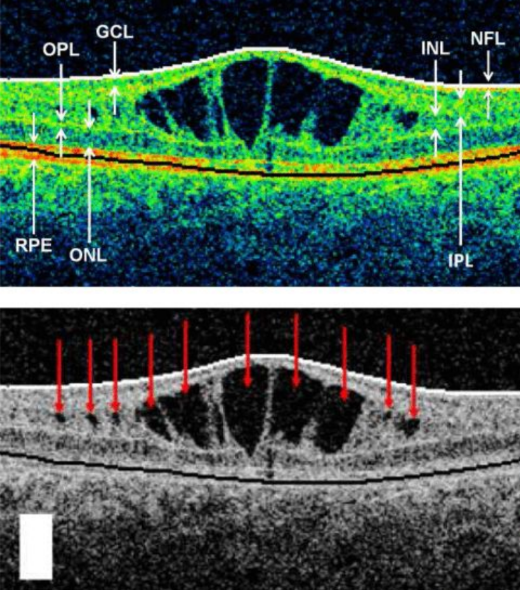}
\caption{Figure(above) represents 7 intra retinal layers with cyst in it. Figure(below) shows true cystic regions in its ground truth image.}
\end{figure}

Before segmentation in OCT images can be successfully achieved, denoising must be performed to mitigate the effects of speckle. Speckle occurs in both OCT and ultrasonic imaging, and arises from the random interference of waves reflected from sub resolution variances within the object. Maintaining edge-like features in the image after speckle denoising is particularly important in segmentation applications.

The brief introduction about disorder and CNN has been described first. Then detailed discussions about problem statement is stated followed by the explanation of the proposed methodology, experimental setup and results. Also comparison methods and comparisons with previous approaches are discussed at the end. The proposed method gives more than 70\% dice coefficient on almost all vendor types.

\subsection{Cystoid Macular Edema (CME)}
Cystoid macular edema or CME, is a painless disorder which affects the central retina or macula. When this condition is present, multiple cyst-like (cystoid) areas of fluid appear in the macula and cause retinal swelling or edema.

The symptoms of CME are Blurred or decreased central vision. Although the exact cause of CME is not known, it may accompany a variety of diseases such as retinal vein occlusion, uveitis, or diabetes. It most commonly occurs after cataract surgery.

About 1-3\% of those who have cataract extractions will experience decreased vision due to CME, usually within a few weeks after surgery. If the disorder appears in one eye, there is an increased risk (possibly as high as 50\%) that it will also affect the second eye. Fortunately, however, most patients recover their vision with observation or treatment.

\subsection{Artificial Neural Network (ANN)}
The inventor of the first neurocomputer, Dr. Robert Hecht-Nielsen, defines a neural network as

"...a computing system made up of a number of simple, highly interconnected processing elements, which process information by their dynamic state response to external inputs."

An Artificial Neural Network (ANN) is an information processing paradigm that is inspired by the way biological nervous systems, such as the brain, process information. The key element of this paradigm is the novel structure of the information processing system. It is composed of a large number of highly interconnected processing elements (neurons) working in unison to solve specific problems [4]. ANNs, like people, learn by example. An ANN is configured for a specific application, such as pattern recognition or data classification, through a learning process. Learning in biological systems involves adjustments to the synaptic connections that exist between the neurones. This is true of ANNs as well [7].

There are two Artificial Neural Network topologies - FeedForward and Feedback.

\subsubsection{FeedForward ANN}
The information flow is unidirectional. A unit sends information to other unit from which it does not receive any information. There are no feedback loops. They are used in pattern generation/recognition/classification. They have fixed inputs and outputs.

Feed-forward networks have the following characteristics:
1. Perceptrons are arranged in layers, with the first layer taking in inputs and the last layer producing outputs [3]. The middle layers have no connection with the external world, and hence are called hidden layers.
2. Each perceptron in one layer is connected to every perceptron on the next layer. Hence information is constantly "fed forward" from one layer to the next., and this explains why these networks are called feed-forward networks [4].
3. There is no connection among perceptrons in the same layer.

\begin{figure}[htbp]
\centering
\includegraphics[width=\linewidth]{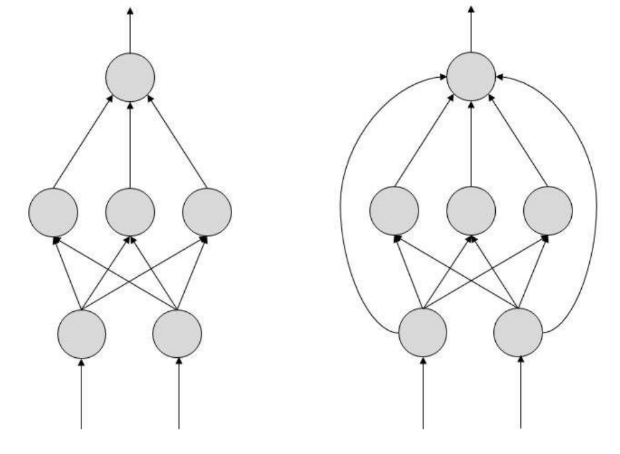}
\caption{Feedforward network}
\end{figure}

\subsubsection{FeedBack ANN}
Feedback loops are allowed. They are used in content addressable memories. It is a class of artificial neural network where connections between units form a directed cycle. This creates an internal state of the network which allows it to exhibit dynamic temporal behavior. Unlike feedforward neural networks, RNNs can use their internal memory to process arbitrary sequences of inputs [9]. This makes them applicable to tasks such as unsegmented connected handwriting recognition or speech recognition.

The human brain is a recurrent neural network (RNN): a network of neurons with feedback connections. It can learn many behaviors / sequence processing tasks / algorithms/programs that are not learnable by traditional machine learning methods [5]. This explains the rapidly growing interest in artificial RNNs for technical applications: general computers which can learn algorithms to map input sequences to output sequences, with or without a teacher.

They are computationally more powerful and biologically more plausible than other adaptive approaches such as Hidden Markov Models (no continuous internal states), feedforward networks and Support Vector Machines (no internal states at all). Recent applications include adaptive robotics and control, handwriting recognition, speech recognition, keyword spotting, music composition, attentive vision, protein analysis, stock market prediction, and many other sequence problems [8].

\begin{figure}[htbp]
\centering
\includegraphics[width=\linewidth]{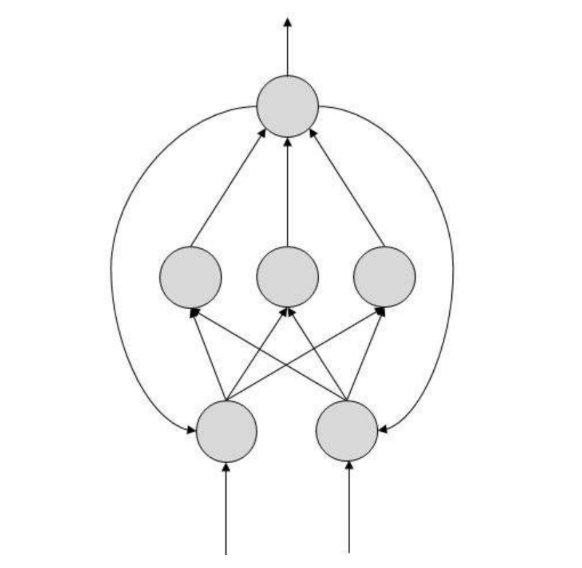}
\caption{Feedback network.}
\end{figure}

\subsubsection{Working of ANNs}
In the topology diagrams shown, each arrow represents a connection between two neurons and indicates the pathway for the flow of information. Each connection has a weight, an integer number that controls the signal between the two neurons [5].

If the network generates a "good or desired" output, there is no need to adjust the weights. However, if the network generates a "poor or undesired" output or an error, then the system alters the weights in order to improve subsequent results.

\subsubsection{Back Propagation Algorithm}
It is the training or learning algorithm. It learns by example. If you submit to the algorithm the example of what you want the network to do, it changes the network's weights so that it can produce desired output for a particular input on finishing the training [10]. The Backpropagation algorithm is a supervised learning method for multilayer feed-forward networks from the field of Artificial Neural Networks.

Feed-forward neural networks are inspired by the information processing of one or more neural cells, called a neuron. The principle of the backpropagation approach is to model a given function by modifying internal weightings of input signals to produce an expected output signal [1]. The system is trained using a supervised learning method, where the error between the system's output and a known expected output is presented to the system and used to modify its internal state.

Technically, the backpropagation algorithm is a method for training the weights in a multilayer feed-forward neural network. As such, it requires a network structure to be defined of one or more layers where one layer is fully connected to the next layer. A standard network structure is one input layer, one hidden layer, and one output layer.

In classification problems, best results are achieved when the network has one neuron in the output layer for each class value. For example, a 2-class or binary classification problem with the class values of A and B [7]. These expected outputs would have to be transformed into binary vectors with one column for each class value. Such as [1, 0] and [0, 1] for A and B respectively. This is called a one hot encoding.

\subsection{Resnet}
Residual networks are one of the hot new ways of thinking about neural networks, ever since they were used to win the ImageNet competition in 2015. ResNets were originally introduced in the paper Deep Residual Learning for Image Recognition by He et. al.

The remarkable thing about the ResNet architecture is just how crazy deep it is. For comparison, the Oxford Visual Geometry Group released a Very Deep Convolutional Network for Large-Scale Visual Recognition, which even has "Very Deep" in the name, and it had either 16 or 19 layers.

ResNet architectures were demonstrated with 50, 101, and even 152 layers. More surprising than decoupling the number of layers of another architecture, the deeper ResNet got, the more its performance grew. It did very well in the 2015 ImageNet competition, and seems to be the best single model out there for object recognition, with most of the 2016 ImageNet models being ensembles of other models [9].

In machine learning, the vanishing gradient problem is a difficulty found in training artificial neural networks with gradient-based learning methods and backpropagation. In such methods, each of the neural network's weights receives an update proportional to the gradient of the error function with respect to the current weight in each iteration of training.

Traditional activation functions such as the hyperbolic tangent function have gradients in the range (-1, 1), and backpropagation computes gradients by the chain rule. This has the effect of multiplying n of these small numbers to compute gradients of the "front" layers in an n-layer network, meaning that the gradient (error signal) decreases exponentially with n and the front layers train very slowly [4].

With the advent of the backpropagation algorithm in the 1970s, many researchers tried to train supervised deep artificial neural networks from scratch, initially with little success. Sepp Hochreiter's diploma thesis of 1991 formally identified the reason for this failure in the "vanishing gradient problem", which not only affects many-layered feedforward networks, but also recurrent neural networks [7]. The latter are trained by unfolding them into very deep feedforward networks, where a new layer is created for each time step of an input sequence processed by the network.

One of the newest and most effective ways to resolve the vanishing gradient problem is with residual neural networks [10]. It was noted prior to ResNets that a deeper network would actually have higher training error than the shallow network. This intuitively can be understood as data disappearing through too many layers of the network, meaning output from a shallow layer was diminished through the greater number of layers in the deeper network, yielding a worse result.

Going with this intuitive hypothesis, it was found by Microsoft Research that splitting a deep network into chunks (say, each chunk is three layers-tweakable) and passing the input into each chunk straight through to the next chunk (along with the residual-output of the chunk minus the input to the chunk that is reintroduced) helped eliminate much of this disappearing signal problem. No extra parameters or changes to the learning algorithm were needed. ResNets yielded lower training error (and test error) than their shallower counterparts simply by reintroducing outputs from shallower layers in the network to compensate for the vanishing data [2].

\subsubsection{Residuals and Information Flows}
First, here is a graphic illustrating the concept of a "residual network".

\begin{figure}[htbp]
\centering
\includegraphics[width=\linewidth]{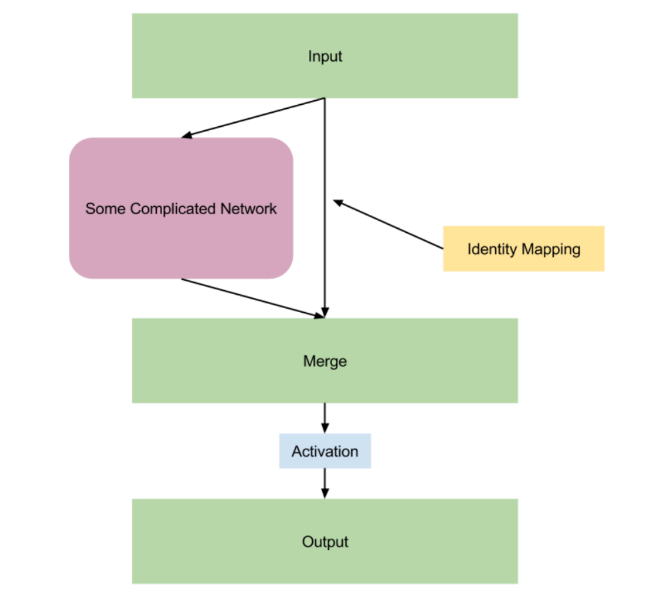}
\caption{Residual Network information flow.}
\end{figure}

The graphic above illustrates how the residual network achieves better information flow by passing more information through identity mapping to avoid going through the residual function.

This "network" acts just as a building block. In the ResNet architecture, the building block repeats many times to form a whole residual network.

\subsubsection{ResNet General Architecture}
Mathematically, we can express this network with input x, some transformation $f(x)$ (the complicated network in the diagram), some merge function $m(x,y)$, some activation function $a(x)$ and output y as

\begin{equation}
y=a(m(x,f(x)))
\end{equation}

Characteristically, $f(x)$ is a convolution or series of convolutions, $m(x,y)$ is simply addition, and $a(x)$ is the rectified linear unit activation function, giving us the equation

\begin{equation}
y=relu(x+f(x))
\end{equation}

The residual is the network error that we want to correct at a particular layer [3].

When people talk about "ResNet" as an abbreviation for "Residual Network", it is usually to refer to convolutional neural networks. But in principle, the idea behind them can be applied to any type of neural network [6]. For example, two researchers at Google recently used residuals applied to a Gated Recurrent Unit for image compression.

\subsubsection{Why are residuals a good idea?}
The central idea of the ResNet paper is that it is a good idea, when adding more layers to a network, to keep the representation more or less the same. In other words, extra layers shouldn't warp the representation very much. Suppose a shallow network perfectly represents the data, and more layers are added. Since the shallow network works perfectly, the best thing for the new layers to do would be to learn the identity function. If the shallow network made a few errors, we would want the new layers to learn to correct the errors, but otherwise not affect the output very much.

Phrased another way, it is an easier learning problem if the network learns to correct the residual error. Once a good representation is learned, the network shouldn't mess with it too much [6]. The other side to this problem is that we want the shallow network to be able to learn a good solution, without having to learn gradients through higher level layers. Phrased yet another way, the residual part should ensure that the representation learned is strictly better than whatever we can get without the residual part.

\subsubsection{Comparison with Highway Networks}
This idea has been phrased differently as information flow, and shows up in LSTM (Long Short Term Memory) and GRU(Gated Recurrent Unit) networks and Highway networks. The key difference between residual and highway networks is the absence of gating [9]. In a highway network, the merge function $m(x,y)$ which for the residual network was simply addition, would instead be expressed as

\begin{equation}
m(x,f(x))=x*g(x)+f(x)*(1-g(x))
\end{equation}

where $g(x)$ is a gating function dependent on x. Depending on how the problem is formulated, the gating function can significantly increase the number of parameters. Aside from that, this formulation might get in the way of the idea discussed earlier; we want the lower layers to learn a near-perfect representation, so we should avoid modifying this representation at all in upper layers.

\begin{figure}[htbp]
\centering
\includegraphics[width=\linewidth]{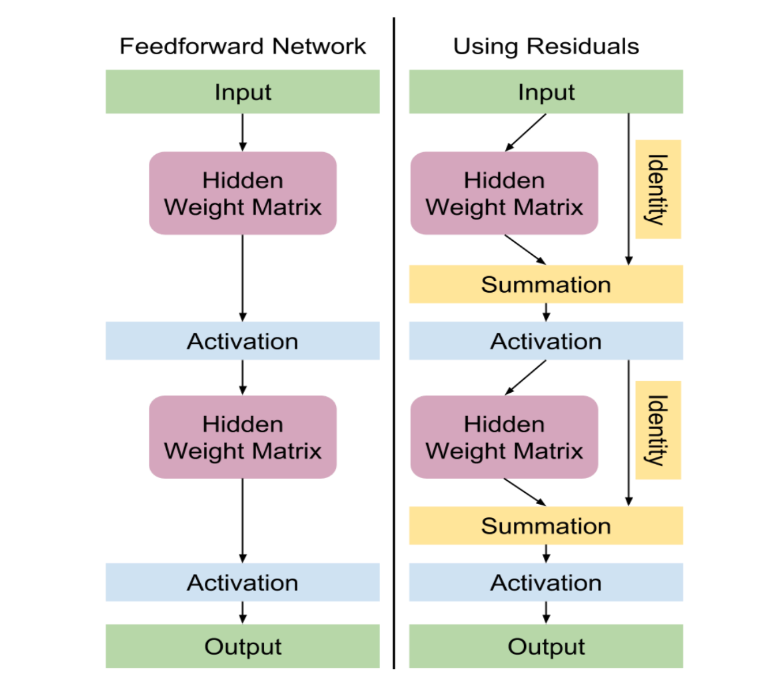}
\caption{Feedforward Network vs Residual Network}
\end{figure}

Allegedly, the key advantage of doing this is so that more information can flow from the output of the first layer to the end result. Let's take advantage of some loose mathematical notation to characterize what we mean by "information flow" [8]. Intuitively, high information flow between two parameters means that changing one parameter significantly affects the other parameter. In other words, the partial derivative (usually called a "gradient") is relatively large.

For both networks:

\begin{equation}
\frac{\partial\vec{y}}{\partial h1}=\frac{\partial h2}{\partial h1}=\frac{\partial sigm}{\partial h1}
\end{equation}

For the first network, the term $\frac{csigm}{\hat{c}h1}$ has to go through the weight matrix W2 [9]. When the output of the first layer is dotted with the weight matrix to get the output, it means that the gradient suddenly depends on all of the parameters in the weight matrix. Mathematically:

\begin{equation}
\frac{\partial\vec{y}}{\partial h1}=\frac{\partial sigm}{\partial dot}\frac{\partial dot}{\partial h1}
\end{equation}

However, for the second network, there are two routes it can go, with one of them avoiding the weight matrix completely:

\begin{equation}
\frac{\partial\vec{y}}{\partial h1}=\frac{\partial sigm}{\partial dot}\frac{\partial dot}{\partial h1}+\frac{\partial sigm}{\partial h1}
\end{equation}

To take the concept of information flow a step further, we can use a rectified linear unit instead of a sigmoid for our activation function [7]. If we use ReLUs and stack multiple layers together, any positive outputs of any layer are passed along completely, which is really good information flow.

\section{Literature Review}
The study of various approaches by distinguishing the approaches into statistical, graph based, denoising based etc, helped us to study the approaches systematically. In this process of study, other macular edema like Diabetic Macular Edema and Microcystic macular edema approaches were also studied which gave us a broader idea.

\subsection{Threshold based models}
\subsubsection{H-minima Transform and non-linear complex diffusion}
The main goal of this work is a presentation of a new algorithm dedicated to cysts detection in Optical Coherence Tomography (OCT) images. Considered in this paper OCT scans concern the eyeball examination in ophthalmology. The proposed approach utilizes different image processing techniques including non-linear complex diffusion and mathematical morphology operations and algorithms, in particular H-Minima Transform, image reconstruction by erosion, geodesic operations etc. The algorithm to confirm its effectiveness - has been verified based on OCT series [5].

Proposed algorithm was evaluated using OCT series with different types, sizes and localizations of cysts [7]. The analysed data set has included over a dozen OCT series wherein about five have concerned the case of macular edema pathology (contained cysts). Each OCT series consisted of 15-75 slices (2D images). Due to the relatively small database presented results should be considered as preliminary. The fully-automated algorithm has been implemented in MATLAB environment and verified based on desktop computer with Core i7-2600 CPU at 3.4 GHz, 16 GB of RAM and Windows-7 64-bit operating system.

Exemplary results from the analysis of the selected slices are presented. Despite aforementioned image properties correct cysts segmentation has been obtained [3]. The algorithm performance is relatively independent of the size and location of the cysts. Through Pre-Processing stage it is also resistant on local noise and low quality of tomograms.

\subsubsection{Delineating Fluid-Filled Region Boundaries}
We evaluate the ability of a deformable model to yield accurate shape descriptions of fluid-filled regions associated with age-related macular degeneration [4]. Calculation of retinal thickness and volume by the current optical coherence tomography (OCT) system includes fluid-filled regions or lesions along with actual retinal tissue. In order to quantify these lesions independently from the retinal tissue, they must be outlined. A deformable model was applied to OCT images of retinas demonstrating cystoids and subretinal fluid spaces [7]. Several implementation issues were addressed in order to choose appropriate parameters. The use of a nonlinear anisotropic diffusion filter to suppress speckle noise while at the same time preserving the edges of the original image was explored. Once the contours of the lesions were outlined, quantitative analysis of the surface area and volume of the lesions was performed. The deformable model could accurately outline fluid-filled regions within the retina. The detection method tested proved effective in capturing the complexity of fluid-filled regions in OCT images [2]. Deformable models combined with nonlinear anisotropic diffusion filtering show promise in the detection of retinal features of interest for diagnosis in clinical OCT images. Thus, fluid-filled region detection may significantly aid in analysis of treatments and diagnosis.

\subsubsection{Segmentation using fast bilateral filter denoising}
Quantitative tools for assessing CME may lead to better metrics for choosing treatment protocols. To address this need, this paper presents a fully automated retinal cyst segmentation technique for OCT image stacks acquired from a commercial scanner. The proposed method includes a computationally fast bilateral filter for speckle denoising while maintaining CME boundaries [7]. The proposed technique was evaluated in images from 16 patients with vitreoretinal disease and three controls. The average sensitivity and specificity for the classification of cystoid regions in CME patients were found to be 91\% and 96\%, respectively, and the retinal volume occupied by cystoid fluid obtained by the algorithm was found to be accurate within a mean and median volume fraction of 1.9\% and 0.8\%, respectively.

This paper presented a method to segment and quantify the total volume occupied by CME from OCT image stacks, in order to provide a metric that can be evaluated as a potential diagnostic for visual acuity. While the average sensitivity, defined by counting individual cystoid ROIs, was 91\%, we found that missed cystoid ROIs were typically small and did not contribute much error to the total volume estimation. The average fractional volume of CME in their sample set of 16 CME patients was 10\%, and average error in fractional volume was 1.9\% when comparing against results by manual inspection.Importantly, the median difference between the cystoid fractional volume by the algorithm and by manual inspection was only 0.8\%. This suggests that, in most patients, this algorithm represents an accurate method of total cystoid volume assessment.

\subsection{Graph Theory models}
\subsubsection{Probability Constrained Graph-Search-Graph-Cut}
An automated method is reported for segmenting 3-D fluid-associated abnormalities in the retina, so-called symptomatic exudate-associated derangements (SEAD), from 3-D OCT retinal images of subjects suffering from exudative age-related macular degeneration. In the first stage of a two-stage approach, retinal layers are segmented, candidate SEAD regions identified, and the retinal OCT image is flattened using a candidate-SEAD aware approach. In the second stage, a probability constrained combined graph search-graph cut method refines the candidate SEADs by integrating the candidate volumes into the graph cut cost function as probability constraints [1]. The proposed method was evaluated on 15 spectral domain OCT images from 15 subjects undergoing intravitreal anti-VEGF injection treatment. Leave-one-out evaluation resulted in a true positive volume fraction (TPVF), false positive volume fraction (FPVF) and relative volume difference ratio (RVDR) of 86.5\%, 1.7\%, and 12.8\%, respectively. The new graph cut-graph search method significantly outperformed both the traditional graph cut and traditional graph search approaches $(p<0.01, p<0.04)$ and has the potential to improve clinical management of patients with choroidal neovascularization due to exudative age-related macular degeneration.

\subsubsection{Graph theory and dynamic programming Stephanie approach}
This paper presents a generalized framework for segmenting closed-contour anatomical and pathological features using graph theory and dynamic programming (GTDP). More specifically, the GTDP method previously developed for quantifying retinal and corneal layer thicknesses is extended to segment objects such as cells and cysts. The presented technique relies on a transform that maps closed-contour features in the Cartesian domain into lines in the quasi-polar domain. The features of interest are then segmented as layers via GTDP. Application of this method to segment closed-contour features in several ophthalmic image types is shown. Quantitative validation experiments for retinal pigmented epithelium cell segmentation in confocal fluorescence microscopy images attests to the accuracy of the presented technique.

In summary, They demonstrated the extension of their automatic GTDP framework to segment not only layered, but also closed-contour structures. Implementation of the algorithm for RPE cell segmentation in confocal fluorescence images of flat-mounted AMD mouse retina resulted in an accurate extraction of cell count and average cell size. We believe that such a tool will be extremely useful for future studies, which use cell morphology as a biomarker for the onset and progression of disease. While validation has yet to be shown for other ophthalmic applications, They have demonstrated preliminary results for intra-retinal cyst segmentation in a SDOCT image as well as cone photoreceptors segmentation in an AOSO image [5]. This is highly encouraging for reducing the time and manpower required for segmenting closed-contour features in ophthalmic studies.

\subsubsection{Random forest classifier}
Optical coherence tomography (OCT) is a powerful imaging tool that is particularly useful for exploring retinal abnormalities in ophthalmological diseases. Recently, it has been used to track changes in the eye associated with neurological diseases such as multiple sclerosis (MS) where certain tissue layer thicknesses have been associated with disease progression. A small percentage of MS patients also exhibit what has been called microcystic macular edema (MME), where fluid collections that are thought to be pseudocysts appear in the inner nuclear layer. Very little is known about the cause of this condition so it is important to be able to identify precisely where these pseudocysts occur within the retina [3]. This identification would be an important first step towards furthering their understanding. In this work, They present a detection algorithm to find these pseudocysts and to report on their spatial distribution. Our approach uses a random forest classifier trained on manual segmentation data to classify each voxel as pseudocyst or not. Despite having a small sample size of five subjects, the algorithm correctly identifies 84.6\% of pseudocysts as compared to manual delineation. Finally, using their method, They show that the spatial distribution of pseudocysts within the macula are generally contained within an annulus around the fovea [1].

\subsubsection{Combining Region Flooding and Texture Analysis}
In this work Optical Coherence Tomography (OCT) retinal images are automatically processed to detect the presence of cysts. The methodology is composed by three phases: region of interest where cysts will be searched is delimited; a watershed algorithm is applied to find all the possible regions in the image which might conform cystic structures; finally, texture analysis is performed in each region from previous phase to final classification. Results show that accuracy achieved with this method is over 80\%.

In this paper, a first approach to an automatic methodology for cyst detection in OCT retinal images has been presented [11]. It is based on using watershed algorithm to detect candidate regions on the image and then, discard all the possible regions to reduce the set of candidates, which because of some of their properties could be considered as cysts. Finally, a classifier determines if they correspond to cystic regions based on texture descriptors extracted from them. Analysis show this is a good approach to this problem, given that results in classifications are around 80\% precision [1].

Other possibilities can be studied to improve the proposed methodology, such as distinguishing categories of cyst based on their location on the image or detecting vessel shades to reduce the number of candidates.

\subsection{Statistical models}
\subsubsection{Kernel regression based segmentation}
We present a fully automatic algorithm to identify fluid-filled regions and seven retinal layers on spectral domain optical coherence tomography images of eyes with diabetic macular edema (DME). To achieve this, they developed a kernel regression (KR)-based classification method to estimate fluid and retinal layer positions. We then used these classification estimates as a guide to more accurately segment the retinal layer boundaries using their previously described graph theory and dynamic programming (GTDP) framework [4]. We validated their algorithm on 110 Bscans from ten patients with severe DME pathology, showing an overall mean Dice coefficient of 0.78 when comparing their KR + GTDP algorithm to an expert grader. This is comparable to the inter-observer Dice coefficient of 0.79. The entire data set is available online, including their automatic and manual segmentation results. To the best of their knowledge, this is the first validated, fully-automated, seven-layer and fluid segmentation method which has been applied to real-world images containing severe DME.

We developed a fully automatic algorithm to identify fluid and eight retinal layer boundaries on SD-OCT images with DME pathology [9]. Results showed that their automatic algorithm performed comparably to manual graders. This is the first validated, fully-automated, and multi-layer segmentation method which has been applied to real-world, clinical images containing severe DME. Accurate identification of DME imaging biomarkers is extremely important, as it will facilitate the quantification and understanding of DME

\subsubsection{Three-Dimensional Analysis of Retinal Layer Texture:}
In this paper, a method for automated characterization of the normal macular appearance in spectral domain OCT (SD-OCT) volumes is reported together with a general approach for local retinal abnormality detection. Ten intra retinal layers are first automatically segmented and the 3-D image dataset flattened to remove motion-based artifacts. From the flattened OCT data, 23 features are extracted in each layer locally to characterize texture and thickness properties across the macula. The normal ranges of layer-specific feature variations have been derived from 13 SD-OCT volumes depicting normal retinas. Abnormalities are then detected by classifying the local differences between the normal appearance and the retinal measures in question [1]. This approach was applied to determine footprints of fluid-filled regions-SEADs (Symptomatic Exudate-Associated Derangements) in 78 SD-OCT volumes from 23 repeatedly imaged patients with choroidal neovascularization (CNV), intra-, and sub-retinal fluid and pigment epithelial detachment. The automated SEAD footprint detection method was validated against an independent standard obtained using an interactive 3-D SEAD segmentation approach. An area under the receiver-operating characteristic curve of was obtained for the classification of vertical, cross-layer, macular columns.A stud performed on 12 pairs of OCT volumes obtained from the same eye on the same day shows that the repeatability of the automated method is comparable to that of the human experts. This work demonstrates that useful 3-D textural information can be extracted from SD-OCT scans and-together with an anatomical atlas of normal retinas can be used for clinically important applications.

\subsection{Miscellaneous models}
\subsubsection{AdaBoost classifier}
An automated method is proposed to segment and quantify the volume of cystoid macular edema (CME) for the abnormal retina with macular hole (MH) in 3D OCT images. The proposed framework consists of three parts: (1) preprocessing, which includes denoising, intraretinal layers segmentation and flattening, MH and vessel silhouettes exclusion; (2) coarse segmentation, in which an AdaBoost classifier is used to get the seeds and constrained regions for Graph Cut; (3) fine segmentation, in which a graph cut algorithm is used to get the refine segmentation result. The proposed method was evaluated in 3D OCT images from 18 typical patients with CMEs and MH. The true positive volume fraction (TPVF), false positive volume fraction (FPVF) and accuracy rate (ACC) for CME volume segmentation are 84.6\%, 1.7\% and 99.7\%, respectively.

\subsubsection{Fuzzy level-set model}
Proposed model is a novel automated volumetric segmentation method to detect and quantify retinal fluid on optical coherence tomography (OCT). The fuzzy level set method was introduced for identifying the boundaries of fluid filled regions on B-scans (x and y-axes) and C-scans (z-axis). The boundaries identified from three types of scans were combined to generate a comprehensive volumetric segmentation of retinal fluid. Then, artefactual fluid regions were removed using morphological characteristics and by identifying vascular shadowing with OCT angiography obtained from the same scan. The accuracy of retinal fluid detection and quantification was evaluated on 10 eyes with diabetic macular edema. Automated segmentation had good agreement with manual segmentation qualitatively and quantitatively [7]. The fluid map can be integrated with OCT angiogram for intuitive clinical evaluation.

We developed an automated volumetric segmentation method to quantify retinal fluid (IRF and SRF) on OCT which involves three main steps: (1) segment and flatten retinal layers; (2) identify retinal fluid space using a fully automated and self-adaptive model (fuzzy level-set method) on OCT cross-sections from three orthogonal directions (two types of B-scans and C-scans); (3) remove remaining artifacts by identifying morphological characteristics and vascular shadowing. We showed that the proposed algorithm can detect and quantify the retinal fluid in DME eyes with a varied image contrasts. The fuzzy level-set algorithm agreed with expert human grading very well. Their technique offers a major advance in providing clinically valuable quantitative measurements of IRF and SRF.

\subsubsection{Pathological Cavities Segmentation}
PURPOSE. To develop and evaluate a method for automated segmentation and quantitative analysis of pathological cavities in the retina visualized by spectral-domain optical coherence tomography (SD-OCT) scans.

METHODS. The algorithm is based on the segmentation of the gray-level intensities within a Bscan by a k-means cluster analysis and subsequent classification by a k-nearest neighbor algorithm. Accuracy was evaluated against three clinical experts using 130 bullous cavities identified on eight SD-OCT B-scans of three patients with wet age-related macular degeneration (AMD) and five patients with X-linked retinoschisis, as well as on one volume scan of a patient with X-linked retinoschisis. The algorithm calculated the surface area of the cavities for the B-scans and the volume of all cavities for the volume scan. In order to validate the applicability of the algorithm in clinical use, we analyzed 31 volume scans taken over the course of 4 years for one AMD patient with a serous retinal detachment [8].

RESULTS. Discrepancies in area measurements between the segmentation results of the algorithm and the experts were within the range of the area deviations among the experts. Volumes interpolated from the B-scan series of the volume scan were comparable among experts and algorithm. Volume changes of the serous retinal detachment were quantifiable.

CONCLUSIONS. The segmentation algorithm represents a method for the automated analysis of large numbers of volume scans during routine diagnostics and in clinical trials.

\subsection{Other Macular Edema models}
\subsubsection{Automated Localization of Cysts in DME}
This paper presents a novel automated system that localizes cysts in optical coherence tomography (OCT) images of patients with diabetic macular edema (DME) [7]. First, in each image, six sub-retinal layers are detected using an iterative high-pass filtering approach. Next, significantly dark regions within the retinal micro-structure are detected as candidate cystoid regions. Each candidate cystoid region is then further analyzed using solidity, mean and maximum pixel value of the negative OCT image as decisive features for estimating the area of cystoid regions. The proposed system achieves 90\% correlation between the estimated cystoid area and the manually marked area, and a mean error of 4.6\%. Finally the proposed algorithm locates the cysts in the inner plexiform region, inner nuclear region and outer nuclear region with an accuracy of 88\%, 86\% and 80\%, respectively.

This paper has proposed an automated diabetic cyst localization system using OCT images of patients with DME [8]. The automated area of cysts detected is highly correlated with the manually marked cysts $(r=0.9)$ and the proposed system over estimates cystoid area such that the mean ratio between the actual cyst area and detected area is 0.85. Also, their system results in a mean error of 4.6\% in estimating the percentage of the area within the region of interest occupied by cysts as compared to existing works that report upto 12\% mean error in cystoid estimation. Finally, their system can detect cysts in the inner plexiform, inner nuclear and outer nuclear regions with accuracy of 88\%, 86\% and 80\%, respectively [3]. Future efforts will be directed towards analysis of additional OCT image stacks and automated estimation of the volume of cystoids from other imaging systems apart from the Spectralis. The proposed automated system can aid clinical studies aimed at monitoring visual prognosis and disease progression.

\subsubsection{Microcystic macular edema in OCT}
Microcystic macular edema (MME) manifests as small, hyporeflective cystic areas within the retina. For reasons that are still largely unknown, a small proportion of patients with multiple sclerosis (MS) develop MME-predominantly in the inner nuclear layer. These cystoid spaces, denoted pseudocysts, can be imaged using optical coherence tomography (OCT) where they appear as small, discrete, low intensity areas with high contrast to the surrounding tissue [11]. The ability to automatically segment these pseudocysts would enable a more detailed study of MME than has been previously possible. Although larger pseudocysts often appear quite clearly in the OCT images, the multi-frame averaging performed by the Spectralis scanner adds a significant amount of variability to the appearance of smaller pseudocysts. Thus, simple segmentation methods only incorporating intensity information do not perform well. In this work, They propose to use a random forest classifier to classify the MME pixels [4]. An assortment of both intensity and spatial features are used to aid the classification. Using a cross-validation evaluation strategy with manual delineation as ground truth, their method is able to correctly identify 79\% of pseudocysts with a precision of 85\%. Finally, They constructed a classifier from the output of their algorithm to distinguish clinically identified MME from non-MME subjects yielding an accuracy of 92\%.

\subsubsection{Segmentation with an exploratory longitudinal study}
Microcystic macular edema (MME) is a term used to describe pseudocystic spaces in the inner nuclear layer (INL) of the human retina [2]. It has been noted in multiple sclerosis (MS) as well as a variety of other diseases. The processes that lead to MME formation and their change over time have yet to be explained sufficiently. The low rate at which MME occurs within such diverse patient groups makes the identification and consistent quantification of this pathology important for developing patient-specific prognoses. MME is observed in optical coherence tomography (OCT) scans of the retina as changes in light reflectivity in a pattern suggestive of fluid accumulations called pseudocysts. Pseudocysts can be readily identified in higher signal-to-noise ratio (SNR) images, however pseudocysts can be indistinguishable from noise in lower SNR scans. In this work, They Expand upon their earlier MME identification methods on Spectralis OCT scans to handle lower quality Cirrus OCT scans. Our approach uses a random forest classifier, trained on manual segmentation of ten subjects, to automatically detect MME. The algorithm has a true positive rate for MME identification of 0.95 and a Dice score of 0.79. We include a preliminary longitudinal study of three patients over four to five years to explore the longitudinal changes of MME. The patients with relapsing-remitting MS and neuromyelitis optica appear to have dynamic pseudocyst volumes, while the MME volume appears stable in the one patient with primary progressive MS.

In this work, an algorithm was created to detect MME within the retina from images acquired with the Zeiss Cirrus OCT machine. The algorithm was validated with a leave-one-out method on ten subjects, and performs well with a TP rate of 95\%. The impressive accuracy of the TP rate is partially due to the removal of all pseudocysts less than 14 pixels in size. They evaluated the performance on an additional set of ten MME and ten HC subjects. The results illustrate a significant difference in the number of pseudocysts in the two cohorts, which suggests that the algorithm can distinguish between MME and non-MME subjects [8]. They also analyzed three subjects longitudinally. The RRMS and NMO subjects exhibit increasing and decreasing trends at different frequencies, while ignoring the baseline of the PPMS subject would suggest a stable MME appearance throughout the time frame of that subject. This work enables improved quantitative analyses of MME. The longitudinal nature and correlation with the disease process underlying the formation of MME can now be further explored on larger scale data sets. A future improvement could include incorporation of spatial distribution of pseudocysts between B-scans, as pseudocysts are present on adjacent scans. Planned future studies include analysis of longitudinal data to more closely understand MME patterns.

\subsection{Problem Definition}
Automatic segmentation and volumetric quantification of intra-retinal cysts from optical coherence tomography scans using vendor independent techniques. The problem can be subdivided into following 3 subproblems
\begin{itemize}
\item Segmentation of retinal layers from OCT scans.
\item Segmentation of intra - retinal cystic fluids from retinal OCT scans.
\item Volumetric quantification of cystic fluid spaces in retina.
\end{itemize}

State of the methods work for high quality images like spectralis OCT. For low quality images the current algorithms work with very poor accuracy. This promotes the need to come up with better and generalized algorithm which works for all vendors or come up with denoising techniques that would produce results of better accuracy.

\section{Proposed Solution}
\subsection{System Requirements}
Software and Hardware Requirements are listed below:

\subsubsection{Hardware Requirements}
\begin{itemize}
\item Device name: Lenovo ThinkStation - D30
\item Processor: Intel\textsuperscript{\textregistered} Xeon\textsuperscript{\textregistered} CPU E-2650 v2 @ 2.60GHz x 16
\item Graphics: Nvidia Quadro K2000/PCCIe/SSE2
\item RAM: 64 GB
\item Disk capacity: 27.0 GB HDD 3.3
\end{itemize}

\subsubsection{Software Requirements}
Ubuntu 14.04 LTS
OS type: 64-bit
Deep Learning Framework:
a. Front-end: Keras
b. Back-end: TensorFlow
Nvidia driver
\begin{itemize}
\item CUDA - parallel computing platform and application programming interface (API) model created by Nvidia.
\item cuDNN - NVIDIA CUDA\textsuperscript{\textregistered} Deep Neural Network library
\end{itemize}

\subsection{Data Description}
OPTIMA cyst challenge dataset: The OPTIMA cyst challenge dataset contains OCT scans with cystoid macular edema obtained using platforms from four different vendors, namely Zeiss Cirrus, Nidek, Spectralis Heidelberg and Topcon [2].

Spectralis vendor OCT scans were used for validation in this study because of better quality and the presence of more features and pathologies (including foveal scans, hard exudates, blood vessel shadows, epiretinal membrane and ILM folds) compared to other vendor scans.

This dataset consists of four OCT volumes, each with 49 frames, acquired over $6\times6$ mm of the macula and foveal center from subjects with CME. All frames provided were gray scale with resolution of $496\times512$ and $496\times1026$ pixels [5].

\begin{table}[htbp]
\centering
\caption{Description of Dataset}
\begin{tabular}{lccccr}
\toprule
Set & Cirrus & Spectralis & Topcon & Nidek & Total \\
\midrule
Training & 4 & 4 & 4 & 4 & 15 \\
Testing 1 & 2 & 2 & 2 & 2 & 8 \\
Testing 2 & 2 & 2 & 2 & 1 & 7 \\
\bottomrule
\end{tabular}
\end{table}

\subsection{Preprocessing}
This stage involves 2 steps preprocessing dataset before feeding into Resnet model.

\subsubsection{Conversion of 3D images to noiseless 2D images}
Obtained 3D volumes are broken down into individual 2D frames. This gives us 1676 training images and 909 testing images.

Ground truth training data is constructed as the union of the two graders to increase the number of positive samples.

Images obtained from Spectralis scanner are 16-bit. Converted them to have pixel ranges between 0-255.

Layer segmentation is performed through the Iowa Reference Algorithm and only the retinal band enclosed by the first 7 layers are considered for further processing. Rest of the image is cropped.

Non-local means filtering applied to denoise the images. Parameters are taken heuristically. Different for every vendor.

Contrast Limited Adaptive Histogram Equalization (CLAHE) is used to enhance the contrast of the images. This accentuates the difference in intensity of the cyst and non-cyst regions.

All images are resized to size 256x512 to feed into the CNN.

\subsubsection{Breaking down 2D images into blocks}
The resnet18 model is used. Basically it is a classification algorithm. And as we need to segment the cyst region and not just detect it we need to segment the cyst region. And hence our approach is segmentation via classification by dividing image into a number of cyst patches. The entire image size is $256\times512$ We divide it into number of $11\times11$ non overlapping patches. As we have around 1600 images, we got around 30000 cyst patches, And we randomly chose equal number of non cyst patches from all vendors prioritizing mainly on the central 50 $(5\times10)$ patches as they will have major impact in segmenting cyst patches.

For label of each patch the center of the patch is considered to classify between cyst and non cyst patches. We trained the model using this from scratch without using any pre trained weights. It took around 3 to 4 days to train.

While testing we need to follow the same steps in reverse order. But this time we need to consider overlapping patches if we need to classify each and every pixel of the image as cyst or non cyst.

\subsection{Training the Model}
The 3D OCT volumes are first broken down into 2D images along the b-scans. This leaves us with 1676 training images and 909 testing images across all vendors. All the images include the choroid and retinal barrier in the images along with the actual retinal layer.As discussed, only the region of interest between the ILM and OPE retinal layers are considered and rest of the image is cropped. All the images are then resized using interpolation to a standard resolution of 256x512 to feed into the network. The network is trained with "LOSS FUNCTION" with a batch size of 4 images. The Adam optimizer is used to update the weights $(\beta1=0.9,\beta2=0.999,=1e-08).$ The learning rate was set to be 3e-4, found to give the best results with a random search. No pre-trained weights were used in the process. The network was trained completely on the specified training set. To account for the relatively less number of training samples, heavy data augmentation was applied during training. New data is generated by transforming the original data by horizontal flips and random rotations, height, width and zoom shifts. These transformations are applied "on-the-fly" during the training time, alleviating storage concerns.

\subsubsection{Model parameters summary}
Following were the parameters used while training our model.
Learning rate decided to be 3e-4.
Optimizer used: Adam.
Batch Normalization added to the network.
Loss function: categorical cross-entropy.
Number of epochs: 100
Time taken for training: 3 to 4 days.

\begin{figure}[htbp]
\centering
\includegraphics[width=\linewidth]{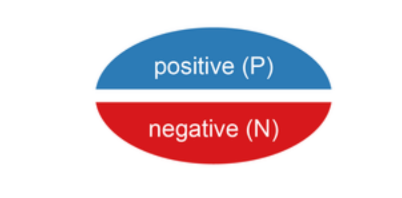}
\caption{Two actual classes or observed labels}
\end{figure}

\section{Experiment Result and Discussion}
\subsection{Comparative measures}
The various comparative measures used to check the accuracy of the proposed approach is explained below.

Test dataset for evaluation
A dataset used for performance evaluation is called a test dataset. It should contain the correct labels (observed labels) for all data instances. These observed labels are used to compare with the predicted labels for performance evaluation after classification.

In binary classification, a test dataset has two labels; positive and negative.

\begin{figure}[htbp]
\centering
\includegraphics[width=\linewidth]{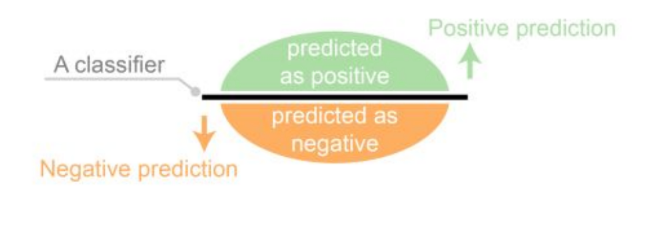}
\caption{Predicted classes of a perfect classifier}
\end{figure}

Predictions on test datasets
The predicted labels will be exactly the same if the performance of a binary classifier is perfect, but it is uncommon to be able to develop a perfect binary classifier that is practical for various conditions [4].

The performance of a binary classifier is perfect when it can predict the exactly same labels in a test dataset [2].

Hence, the predicted labels usually match with part of the observed labels.

\begin{figure}[htbp]
\centering
\includegraphics[width=\linewidth]{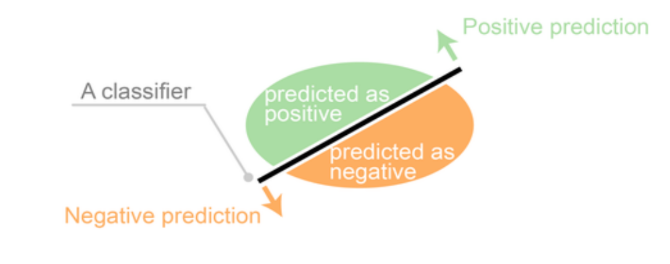}
\caption{Predicted classes of a classifier}
\end{figure}

The predicted labels of a classifier match with part of the observed labels.

\begin{figure}[htbp]
\centering
\includegraphics[width=\linewidth]{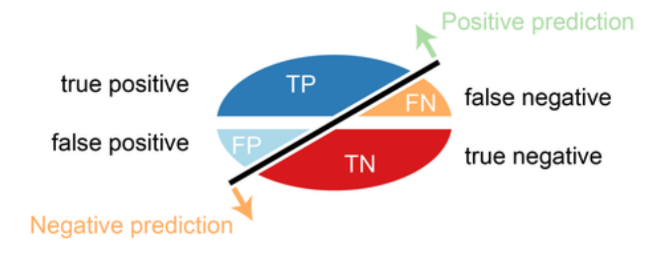}
\caption{Four outcomes of a classifier}
\end{figure}

Confusion matrix from the four outcomes
A confusion matrix is formed from the four outcomes produced as a result of binary classification.

Four outcomes of classification
A binary classifier predicts all data instances of a test dataset as either positive or negative [4]. This classification (or prediction) produces four outcomes true positive, true negative, false positive and false negative.
\begin{itemize}
\item True positive (TP): correct positive prediction
\item False positive (FP): incorrect positive prediction
\item True negative (TN): correct negative prediction
\item False negative (FN): incorrect negative prediction
\end{itemize}

Classification of a test dataset produces four outcomes - true positive, false positive, true negative, and false negative.

\begin{equation}
Dice coefficient=2\frac{|Detected\cap GT|}{|Detected|+|GT|}
\end{equation}

where Detected represents predicted positive values and GT represents positive values marked on Ground truth.

\begin{equation}
Precision=\frac{TP}{TP+FP}
\end{equation}

\begin{equation}
Sensitivity/Recall=\frac{TP}{TP+FN}
\end{equation}

These are the statistical measures used for comparing true and predicted results.

\subsection{Quantitative Analysis of Results obtained for different graders}

\begin{table}[htbp]
\centering
\caption{Test Results for Grader 1}
\begin{tabular}{lccc}
\toprule
Volume & Dice Coefficient & Sensitivity/Recall & Precision \\
\midrule
Cirrus\_1 & 0.8893 & 0.8018 & 0.9982 \\
Cirrus\_2 & 0.8156 & 0.6892 & 0.9986 \\
Cirrus\_3 & 0.917 & 0.8475 & 0.999 \\
Cirrus\_4 & 0.93 & 0.8721 & 0.9961 \\
Mean & 0.888 & 0.8027 & 0.998 \\
Std & 0.0511 & 0.0810 & 0.0013 \\
Nidek 1 & 0.8523 & 0.748 & 0.9904 \\
Nidek 2 & 0.7906 & 0.6681 & 0.9682 \\
Nidek 3 & 0.8221 & 0.7242 & 0.9506 \\
Mean & 0.8217 & 0.7134 & 0.9697 \\
Std & 0.0309 & 0.0410 & 0.0199 \\
Spectralis\_1 & 0.7824 & 0.6438 & 0.997 \\
Spectralis\_2 & 0.8327 & 0.7134 & 0.999 \\
Spectralis\_3 & 0.8484 & 0.7383 & 0.997 \\
Spectralis\_4 & 0.8467 & 0.7391 & 0.991 \\
Mean & 0.8276 & 0.7087 & 0.996 \\
Std & 0.0309 & 0.0448 & 0.0035 \\
Topcon\_1 & 0.8559 & 0.7487 & 0.999 \\
Topcon\_2 & 0.8072 & 0.6791 & 0.995 \\
Topcon\_3 & 0.5463 & 0.3762 & 0.997 \\
Topcon\_4 & 0.8461 & 0.7382 & 0.9909 \\
Mean & 0.7639 & 0.6356 & 0.9955 \\
Std & 0.1466 & 0.1756 & 0.0035 \\
Mean & 0.8255 & 0.7152 & 0.9911 \\
Std & 0.0878 & 0.7097 & 0.9894 \\
\bottomrule
\end{tabular}
\end{table}

\begin{table}[htbp]
\centering
\caption{Test Results for Grader 2}
\begin{tabular}{lccc}
\toprule
Volume & Dice Coefficient & Sensitivity/Recall & Precision \\
\midrule
Cirrus 1 & 0.8893 & 0.8019 & 0.9981 \\
Cirrus\_2 & 0.8155 & 0.6892 & 0.9986 \\
Cirrus\_3 & 0.9171 & 0.8476 & 0.999 \\
Cirrus\_4 & 0.9302 & 0.8726 & 0.996 \\
Mean & 0.8880 & 0.8028 & 0.9979 \\
Std & 0.0513 & 0.0812 & 0.0013 \\
Nidek 1 & 0.8529 & 0.7489 & 0.9904 \\
Nidek 2 & 0.7922 & 0.6686 & 0.9718 \\
Nidek\_3 & 0.8204 & 0.7197 & 0.9539 \\
Mean & 0.8218 & 0.7124 & 0.9720 \\
Std & 0.0304 & 0.0406 & 0.0183 \\
Spectralis\_1 & 0.7824 & 0.6439 & 0.9969 \\
Spectralis\_2 & 0.8327 & 0.7134 & 0.999 \\
Spectralis\_3 & 0.8479 & 0.7381 & 0.9962 \\
Spectralis\_4 & 0.8464 & 0.7383 & 0.9915 \\
Mean & 0.8274 & 0.7084 & 0.9959 \\
Std & 0.0307 & 0.0446 & 0.0032 \\
Topcon\_1 & 0.8557 & 0.7483 & 0.9992 \\
Topcon\_2 & 0.8072 & 0.679 & 0.9951 \\
Topcon\_3 & 0.5459 & 0.3761 & 0.9956 \\
Topcon\_4 & 0.8455 & 0.7372 & 0.9909 \\
Mean & 0.7636 & 0.6352 & 0.9952 \\
Std & 0.1466 & 0.1754 & 0.0034 \\
Mean & 0.8254 & 0.7149 & 0.9915 \\
Std & 0.8201 & 0.7092 & 0.9899 \\
\bottomrule
\end{tabular}
\end{table}

\begin{table}[htbp]
\centering
\caption{Test Results for Grader 3 (Intersection of Grader 1 \& 2)}
\begin{tabular}{lccc}
\toprule
Volume & Dice Coefficient & Sensitivity/Recall & Precision \\
\midrule
Cirrus 1 & 0.8893 & 0.802 & 0.9979 \\
Cirrus\_2 & 0.8154 & 0.6892 & 0.9981 \\
Cirrus\_3 & 0.9171 & 0.8477 & 0.9988 \\
Cirrus\_4 & 0.9305 & 0.8733 & 0.9957 \\
Mean & 0.8880 & 0.8031 & 0.9976 \\
Std & 0.0514 & 0.0814 & 0.0013 \\
Nidek 1 & 0.854 & 0.7507 & 0.9901 \\
Nidek 2 & 0.7969 & 0.6778 & 0.9667 \\
Nidek\_3 & 0.8261 & 0.7321 & 0.9478 \\
Mean & 0.8257 & 0.7202 & 0.9682 \\
Std & 0.0286 & 0.0379 & 0.0212 \\
Spectralis\_1 & 0.7826 & 0.6442 & 0.9968 \\
Spectralis\_2 & 0.8327 & 0.7135 & 0.999 \\
Spectralis\_3 & 0.8488 & 0.739 & 0.9957 \\
Spectralis\_4 & 0.8475 & 0.741 & 0.9896 \\
Mean & 0.8279 & 0.7094 & 0.9953 \\
Std & 0.0311 & 0.0452 & 0.0040 \\
Topcon\_1 & 0.8561 & 0.7491 & 0.9988 \\
Topcon\_2 & 0.8079 & 0.6804 & 0.9943 \\
Topcon\_3 & 0.5462 & 0.3764 & 0.9952 \\
Topcon\_4 & 0.8467 & 0.7392 & 0.9907 \\
Mean & 0.7642 & 0.6363 & 0.9948 \\
Std & 0.1468 & 0.1759 & 0.0033 \\
Mean & 0.8265 & 0.7170 & 0.9903 \\
Std & 0.8214 & 0.7119 & 0.9885 \\
\bottomrule
\end{tabular}
\end{table}

\subsection{Comparative Study}
The proposed approach is outperforming state of the art methods which is shown in the below table.

\begin{table}[htbp]
\centering
\caption{Results of various State of art methods compared with the Proposed Approach}
\begin{tabular}{lcc}
\toprule
Team & Mean Dice Coeff & Std. Deviation \\
\midrule
de Sisternes et al & 0.68 & 0.14 \\
Venhuizen et al. & 0.601 & 0.18 \\
Oguz et al. & 0.596 & 0.14 \\
Esmaeili et al. & 0.55 & 0.24 \\
Haritz et al. & 0.23 & 0.15 \\
*Proposed approach & 0.82 & 0.08 \\
\bottomrule
\end{tabular}
\end{table}

\section{Conclusion and Future Work}

In literature, several methods have been proposed for the automatic segmentation of intraretinal cysts, a critical task given the severity of cystoid macular edema (CME) and its impact on vision. Historically, progress in this area has been limited by image quality, with previous state-of-the-art approaches achieving a Dice coefficient of merely 68\% and struggling significantly with high-noise scans like those from Topcon. To address this, our work proposed a ResNet Convolutional Neural Network (CNN) approach utilizing patch-wise classification. By evaluating our method on the first publicly available novel cyst segmentation challenge dataset—tested against ground truths from two independent graders across four different vendors—we demonstrated remarkable robustness. Ultimately, our proposed approach achieves an accuracy of over 80\%, outperforming previous state-of-the-art methods across all vendor types, irrespective of underlying image quality. This precise volumetric quantification serves as a vital indicator of CME severity, directly guiding ophthalmologists in clinical treatment planning.

Despite achieving high accuracy, there remains substantial scope for technical improvement. Future work will explore advanced denoising techniques and hyperparameter optimization to reduce time and space complexities during training. Experimenting with alternate block sizes, such as $21\times21$ or $15\times15$, and leveraging pre-trained weights (e.g., ImageNet) could further streamline the training period. Additionally, we plan to evaluate architectures like GoogleNet and U-Net to push segmentation accuracy even higher. Moving beyond 2D patch processing, future iterations will focus on autoregressive or generative models capable of synthesizing and tracking cyst progression over time (3D+Time), borrowing advanced techniques from temporal video synthesis and generative media modeling [17]. Furthermore, as our method transitions to clinical application, ensuring absolute factuality and mitigating false positives ('hallucinations') will be a primary focus, addressing a significant hurdle currently being extensively studied across broader generalized AI models [18].

Beyond core algorithmic enhancements, integrating this tool into real-world clinical workflows presents its own set of challenges and opportunities. To make this diagnostic tool accessible to clinical staff, future development could involve creating a Domain Specific Language (DSL) or a tailored graphical interface, a methodology proven effective in streamlining complex interactive systems such as gaming applications [13]. Furthermore, this segmentation module could eventually act as a foundational node within a larger multi-agent clinical system. Similar to how agentic information retrieval [19] and multi-agent recommenders [16] parse vast amounts of media to suggest relevant content, a medical equivalent could autonomously retrieve historical patient cases and recommend tailored treatment trajectories based on the segmented cyst volume. The robustness of the convolutional and classification approaches demonstrated here can also be extended to other complex computer vision tasks, much like how optimized detection frameworks have been successfully applied to facial expression analysis [14]. Finally, as medical datasets scale, ensuring the privacy and security of patient OCT scans will become paramount. Future iterations of this overarching system could integrate secure data-hiding techniques—similar to semantic steganography used in other textual and numerical domains [15]—to protect electronic health records comprehensively.

\end{document}